\documentclass[conference]{IEEEtran}
\IEEEoverridecommandlockouts
\usepackage{cite}
\usepackage{amsmath,amssymb,amsfonts}
\usepackage{algorithmic}
\usepackage{graphicx}
\usepackage{textcomp}
\usepackage{xcolor}

\usepackage{multirow}
\usepackage{stfloats}
\usepackage{booktabs}
\usepackage{subfigure}
\usepackage{CJKutf8}
\usepackage{hyperref}

\def\BibTeX{{\rm B\kern-.05em{\sc i\kern-.025em b}\kern-.08em
    T\kern-.1667em\lower.7ex\hbox{E}\kern-.125emX}}
\begin{document}

\title{Improving Zero-Shot Chinese-English Code-Switching ASR with $k$NN-CTC and Gated Monolingual Datastores
\thanks{$^{*}$Corresponding author. This work has been supported by the National Key R$\&$D Program of China (Grant No.2022ZD0116307) and NSF China (Grant No.62271270)}
}

\author{\IEEEauthorblockN{Jiaming Zhou$^1$, Shiwan Zhao$^1$, Hui Wang$^1$, Tian-Hao Zhang$^2$, Haoqin Sun$^1$, Xuechen Wang$^1$ and Yong Qin$^{1,*}$} 
\IEEEauthorblockA{$^1$TMCC, College of Computer Science, Nankai University, Tianjin, China\\
$^2$University of Science and Technology Beijing, Beijing, China\\
Email: zhoujiaming@mail.nankai.edu.cn}
}
\maketitle
\begin{abstract}
The $k$NN-CTC model has proven to be effective for monolingual automatic speech recognition (ASR). However, its direct application to multilingual scenarios like code-switching, presents challenges. Although there is potential for performance improvement, a $k$NN-CTC model utilizing a single bilingual datastore can inadvertently introduce undesirable noise from the alternative language. To address this, we propose a novel $k$NN-CTC-based code-switching ASR (CS-ASR) framework that employs dual monolingual datastores and a gated datastore selection mechanism to reduce noise interference. Our method selects the appropriate datastore for decoding each frame, ensuring the injection of language-specific information into the ASR process. We apply this framework to cutting-edge CTC-based models, developing an advanced CS-ASR system. Extensive experiments demonstrate the remarkable effectiveness of our gated datastore mechanism in enhancing the performance of zero-shot Chinese-English CS-ASR.
\end{abstract}


\section{Introduction}
In today's increasingly interconnected world, the ability to accurately transcribe and understand speech in multilingual and code-switching (CS) environments is of paramount importance \cite{moyer2002bilingual}. Automatic speech recognition (ASR) systems play a crucial role in facilitating communication across linguistic boundaries. Recent years have witnessed rapid advancements in ASR technology \cite{prabhavalkar2023end,cif-t}. However, traditional ASR models typically rely on vast amounts of labeled data for each language or dialect, posing a significant challenge in scenarios where such data is scarce or unavailable \cite{wang2024enhancing,slt-wsy, madi,childmandarin, stutter, zhou2024m2r}. Consider, for instance, a scenario where speakers seamlessly transition between two languages, such as English and Chinese, within the same conversation. Traditional ASR systems trained on monolingual datasets struggle to accurately transcribe this CS speech, as similar pronunciations across languages could confuse the model.

Numerous studies have advanced CS-ASR.
In terms of data augmentation, Chi et al. \cite{chi23_interspeech} introduce a method to generate CS text by forcing a multilingual machine translation system to produce CS translations. 
Liang et al. \cite{liang23b_interspeech} propose a novel data augmentation method employing a text-based speech-edit model to improve CS and name entity recognition in ASR.
Other studies \cite{watanabe2017language,8462180,zhang22da_interspeech, Zhou_Li_Sun_Liu_2022} show that jointly modeling ASR and language identity can endow models with some degree of CS capability. However, CS introduces language boundary ambiguity, which can impair the model's language recognition ability \cite{chen2023ba,fan2023language}, leading to performance degradation. To address this issue, Chen et al. \cite{chen2023ba} employ a boundary-aware predictor to acquire representations specifically designed to handle such ambiguity. 

The Mixture of Experts (MoE) technique has also been employed to improve CS-ASR systems. For instance, 
Lu et al. \cite{lu2020bi} introduce a bi-encoder transformer network with MoE architecture to optimize data utilization. Their approach involves the separation of Chinese and English modeling using two distinct encoders to capture language-specific features effectively. Additionally, they employ a gating network to explicitly manage the language identification task. Tan et al. \cite{tan23c_interspeech} propose a lightweight switch routing network to further refine the network. 
However, these methods still require fine-tuning with labeled CS data, a significant challenge due to the impracticality of covering every language pair.
 
Addressing this, researchers have explored zero-shot learning to enable model generalization to CS-ASR task without specific training data. 
Peng et al. \cite{peng2023prompting} propose a prompt engineering method that enhances the Whisper model \cite{whisper} for CS-ASR tasks by replacing a single language token in the prompt with two language tokens.
Yan et al. \cite{yan2023towards} suggest a cross-lingual pseudo-labeling modification for monolingual modules to produce transliterations of foreign speech, aiming to circumvent the error propagation of frame-wise language identification (LID) decisions. Despite this advancement, their approach necessitates fine-tuning each monolingual model with cross-lingual pseudo-labeling and using dual encoders during decoding, highlighting ongoing challenges in CS-ASR development.

Recently, retrieval-augmented methods have gained significant traction in natural language processing (NLP) tasks, such as $k$NN-LM \cite{knnlm} for language modeling and $k$NN-MT \cite{Khandelwal_Fan_Jurafsky_Zettlemoyer_Lewis_2020} for machine translation, proving particularly valuable in managing low-resource scenarios. This technique has also been extensively adopted in speech processing, with numerous studies \cite{knn-ctc,chan2023domain,wang23r_interspeech} leveraging it. Inspired by $k$NN-LM, Zhou et al. \cite{knn-ctc} introduce $k$NN-CTC to improve pre-trained CTC-based ASR systems \cite{ctc} by integrating a $k$NN model to retrieve CTC pseudo labels from a meticulously pruned datastore. Despite its benefits, its application is confined to monolingual settings due to reliance on a single datastore. For zero-shot CS-ASR, adopting $k$NN-CTC with a bilingual Chinese-English datastore offers potential improvements but also poses the risk of decoding interference from extraneous language noise.

In this paper, we concentrate on zero-shot Chinese-English CS-ASR. Leveraging the $k$NN-CTC concept, we introduce a novel approach for zero-shot Chinese-English CS-ASR. Rather than using a combined datastore of both languages, which introduces undesirable noise, we utilize two separate monolingual datastores. Our method features a gated datastore mechanism for selecting the appropriate monolingual datastore for each frame during decoding, thus ensuring the explicit injection of language-specific information.

Our main contributions are as follows:

1. We initially adapt $k$NN-CTC for zero-shot Chinese-English CS-ASR by developing a bilingual datastore, thereby enhancing performance.

2. We then devise a $k$NN-CTC framework that leverages two separate monolingual datastores and implements a selection mechanism to choose the appropriate datastore during decoding, ensuring the precise utilization of language-specific information in conjunction with CTC processing.

3. We demonstrate the effectiveness of our approach through comprehensive experimental validation.

\section{Our method}

\begin{figure}[!t]
  \centering
  \includegraphics[width=\linewidth]{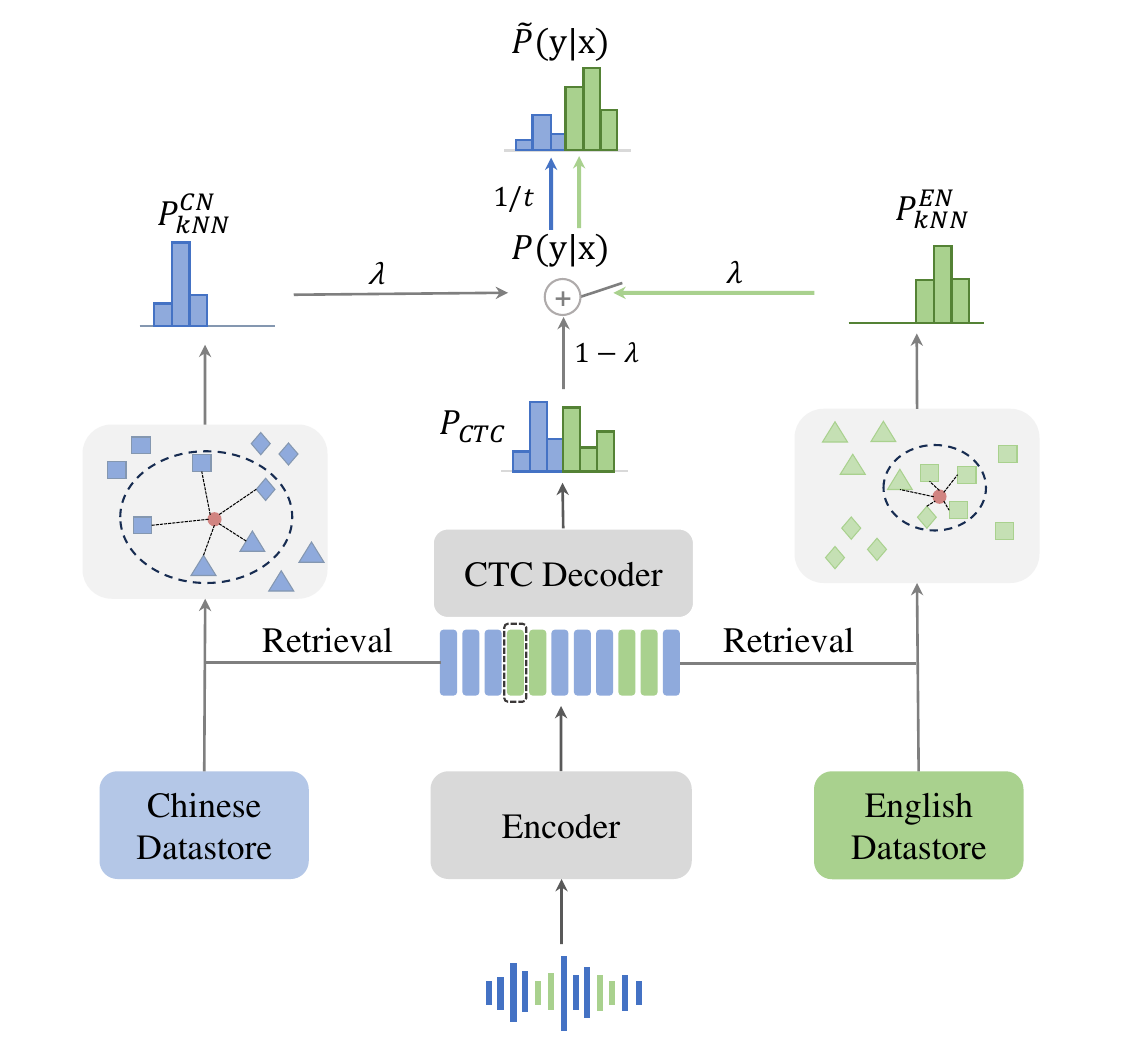}
  \caption{Overview of our methodology employing dual monolingual datastores, with the color blue representing Chinese and green representing English. For each audio frame, two retrieval operations are conducted to identify the appropriate datastore. Following this, the CTC distribution is interpolated with the $k$NN distribution from the selected language (e.g., $P^{EN}_{kNN}$ for English), while the CTC distribution corresponding to the unselected language (in this case, Chinese) is diminished.}
  \label{overview}
\end{figure}

Figure \ref{overview} provides an overview of our proposed methodology. This section will detail our approach to CS-ASR utilizing the $k$NN-CTC model. It will be followed by a comprehensive explanation of our novel implementation of $k$NN-CTC with gated monolingual datastores, detailing each step of the process.

\subsection{CS-ASR based on $k$NN-CTC}

In this section, we introduce how to build a CS-ASR baseline using $k$NN-CTC with a bilingual datastore, comprising two main stages: datastore construction and candidate retrieval. 

\textbf{Datastore construction}:
Given a pre-trained CTC-based ASR model, we first fine-tune the model with the Chinese-only labeled data $S_{CN}$ and English-only labeled data $S_{EN}$.
Subsequently, the CTC pseudo label $\hat{Y}_i$ for the $i$-th frame $X_i$ could be obtained using the equation:
\begin{equation}
    \hat{Y}_i=arg \mathop{max}\limits_{Y_i}P_{CTC}(Y_i|X_i).
    \label{ctc_loss}
\end{equation}
We then extract the intermediate representation $f(X)$ of input $X$ and then employ the CTC pseudo label to create frame-level key-value pairs.
To achieve CS-ASR, we directly construct a single bilingual datastore containing both language datasets. We then construct the datastore by:
\begin{equation}
    D_{ALL}=\{(f(X_i),\hat{Y}_i)|X_i\in S_{CN} \cup S_{EN}\}.
\end{equation}

\textbf{Candidate retrieval}: During decoding, we extract the intermediate representation $f(x)$ from the encoder as a query to retrieve $k$-nearest neighbors $\mathcal{N}_{ALL}$. 
The $k$NN distribution over neighbors aggregates the probability of each vocabulary unit as follows:
\begin{equation}
P_{kNN}(y|x) \propto \sum_{(k_i, v_i)\in\mathcal{N}_{ALL},v_i=y} {exp(-d(k_i,f(x)/\tau))},
\end{equation}
where $\tau$ represents the temperature, $d(\cdot,\cdot)$ is the $L^2$ distance. Subsequently, we interpolate the CTC distribution $P_{CTC}$ with $P_{KNN}$. The final distribution $P(y|x)$ is derived by:
\begin{equation}
\label{final_distribution}
P(y|x)=\lambda P_{kNN}(y|x)+(1-\lambda)P_{CTC}(y|x),
\end{equation}
where $\lambda$ is a hyperparameter to control the weight of $k$NN and CTC.

\subsection{$k$NN-CTC with gated monolingual datastores}
Simply constructing a bilingual datastore may introduce unexpected noise from the alternative language. To address this, we build dual monolingual datastores, noted as $D_{CN}$ for Chinese and $D_{EN}$ for English by the following equation:
\begin{equation}
    D_{CN}=\{(f(X_i),\hat{Y}_i)|X_i\in S_{CN} \},
\end{equation}
\begin{equation}
    D_{EN}=\{(f(X_i),\hat{Y}_i)|X_i\in S_{EN}\}.
\end{equation}

For each input frame $X_i$, we independently retrieve two lists of the $k$ nearest neighbors from the dual datastores. We then calculate the average distances of the  top-$n$ ($n \leq k$) neighbors, denoted as $d_{CN}$ for Chinese and $d_{EN}$ for English, respectively. Setting $n$  to 1 corresponds to the shortest distances among the retrieved neighbors. The injection of language-specific information is determined based on these average distances:
\begin{equation}
    \mathcal{N}_{C} = \begin{cases} 
    \mathcal{N}_{CN}, & \text{if } d_{CN} \leq d_{EN} \\
    \mathcal{N}_{EN}, & \text{otherwise} 
    \end{cases}
\end{equation}
where $\mathcal{N}_{C}$ represents the retrieved $k$ neighbors of the selected language, chosen from either ${N}_{CN}$ or ${N}_{EN}$.
The subsequent step involves deriving the selected monolingual $k$NN distribution from ${N}_{C}$, which is either ${N}_{CN}$ or ${N}_{EN}$, as follows:
\begin{equation}
\label{lang inject}
P_{kNN}(y|x) \propto \sum_{(k_i, v_i)\in\mathcal{N}_{C},v_i=y} {exp(-d(k_i,f(x)/\tau))}.
\end{equation}
The final distribution $P(y|x)$ is obtained as described in Equation \ref{final_distribution}. With the help of language identification-based datastore selection, language-specific information is explicitly injected into the final distribution.

To fully use the language-specific information, we adjust the distribution associated with the alternate language, thereby directly facilitating the determination of the language to which the current frame belongs. Specifically, if the frame is inferred to belong to one language, we reduce the distribution corresponding to the alternate language in the following manner:
\begin{equation}
\label{scale}
    \Tilde{P}(y|x)=
    \begin{cases} 
    P_{CN}/t+P_{EN}, & if \ \mathcal{N}_{C} \ is \ \mathcal{N}_{EN} \\
    P_{CN}+P_{EN}/t, & \text{otherwise} 
    \end{cases}
\end{equation}
where $P_{CN}$ and $P_{EN}$ represent the distributions for Chinese and English within $P(y|x)$, with $t$ serving as the scale temperature to adjust these distributions.

\section{Experimental setup}
\subsection{Dataset}
We utilize ASCEND \cite{lovenia2022ascend}, a Chinese-English dataset for CS-ASR. We split the training set of ASCEND into three parts: Chinese, English, and Mixed (CS-ASR data), denoted as $S_{CN}$, $S_{EN}$, $S_{MIX}$ respectively. 
Our experiments only employ the Chinese and English subsets for fine-tuning the models and constructing datastores to keep the zero-shot setting. 
Additionally, we utilize the ASCEND test set (denoted as TEST) and the mixed training set ($S_{MIX}$) as our test sets. The details of each subset are shown in Table \ref{dataset}.

\begin{table}[]
  \caption{Details of splitting the dataset.}
  \label{dataset}
  \centering
\begin{tabular}{cccc}
\toprule
                       & Subset & \# Utterance & Duration (hr) \\
\midrule
\multirow{2}{*}{Train} & $S_{CN}$  & 4799       & 3.50          \\
                       & $S_{EN}$  & 2331       & 1.65         \\
\midrule
Dev                    & $DEV$     & 1130       & 0.92         \\
\midrule
\multirow{2}{*}{Test}  & $TEST$    & 1315       & 0.92         \\
                       & $S_{MIX}$   & 2739       & 3.62      \\
\bottomrule
\end{tabular}
\end{table}

\begin{table}[]
  \caption{MER (\%) of our proposed method based on the Conformer fine-tuned with $S_{CN}$ and $S_{EN}$.}
  \label{confomer_res}
  \centering
\begin{tabular}{ccccc}
\toprule
    Method     & Datastore             & $TEST$  & $S_{MIX}$ &RTF\\
\midrule
    CTC      & -                     & 26.17 & 28.82 &0.0139\\ 
    $k$NN-CTC  & $D_{ALL}$       & 25.66 & 27.94 &0.0144\\
    Ours    & $D_{CN}$,$D_{EN}$ & \textbf{25.02} & \textbf{26.68} &0.0151\\
\bottomrule
\end{tabular}
\end{table}

\begin{table}[t]
  \caption{MER (\%) of our proposed method based on the Wav2vec2-XLSR fine-tuned with $S_{CN}$ and $S_{EN}$.}
  \label{wav2vec2_res}
  \centering
\begin{tabular}{ccccc}
\toprule
    Method      & Datastore             & $TEST$  & $S_{MIX}$ &RTF\\
\midrule
    CTC      & -                     & 32.65 & 35.22 &0.0123\\
    $k$NN-CTC  & $D_{ALL}$       & 32.47 & 34.94 &0.0133\\
    Ours     & $D_{CN}$,$D_{EN}$ & \textbf{31.48}&  \textbf{33.41} &0.0137\\
\bottomrule
\end{tabular}
\end{table}

\subsection{Implementation details}
Our experiments are conducted using the open-source toolkit WeNet \cite{zhang2022wenet} for Conformer \cite{conformer} and HuggingFace's Transformers \cite{wolf2019huggingface} for Wav2vec2-XLSR \cite{xlsr}. 
We utilize the open-source checkpoint\footnote{ \url{https://github.com/wenet-e2e/wenet/blob/main/docs/pretrained\_models.en.md}}  pre-trained by WenetSpeech \cite{zhang2022wenetspeech}, comprising 12 encoder layers of Conformer and 3 decoder layers of bi-transformer. 
During finetuning, a learning rate of $5 \times 10^{-5}$ and a batch size of 16 with 5000 warmup steps are employed.
For Wav2vec2-XLSR baseline, we leverage the open-source checkpoint\footnote{\url{https://huggingface.co/jonatasgrosman/wav2vec2-large-xlsr-53-chinese-zh-cn}} and fine-tune it with codebase\footnote{\url{https://github.com/HLTCHKUST/ASCEND}} provided by ASCEND \cite{lovenia2022ascend}. We abbreviate Conformer-CTC to Conformer and Wav2vec2-XLSR-CTC to Wav2vec2-XLSR for simplicity. All reported results are derived using CTC greedy search decoding.
It is important to note that our models are fine-tuned exclusively on the subsets $S_{CN}$ and $S_{EN}$, thus preserving the zero-shot CS-ASR setting.

We reimplement the version of $k$NN-CTC (full) \cite{knn-ctc} using FAISS \cite{FAISS}. For Conformer, we follow $k$NN-CTC to determine the location of keys.
For Wav2vec2-XLSR, we select the encoder output as the location of keys. We set $k$ to 1024 following $k$NN-CTC for both baselines. 
$\lambda$ is approximately 0.25 adjusted using the validation set.
Regarding the gated monolingual datastore selection mechanism, we set $n$=300 for the Conformer baseline and $n$=10 for the Wav2vec2-XLSR baseline to compute the average distances of the $n$ nearest neighbors from $D_{CN}$ and $D_{EN}$. We set the distribution calibration temperature $t$ to 5 for the Conformer baseline and 200 for the Wav2vec2-XLSR baseline, adjusting the probability distribution for the alternative language. 
We adopt the Mixture Error Rate (MER) as outlined in \cite{shi2020asru} for metrics. MER accounts for both Chinese characters and English words when calculating the edit distance. Errors are tallied separately for Chinese and English based on the language of the reference token.

\section{Results}
\subsection{Zero-shot CS-ASR results}

The evaluation results for the Conformer and Wav2vec2-XLSR baselines are summarized in Table \ref{confomer_res} and Table \ref{wav2vec2_res}, respectively. Our approach, employing the $k$NN-CTC with a single datastore $D_{ALL}$, as well as our method utilizing dual monolingual datastores $D_{CN}$ and $D_{EN}$, outperform the fine-tuned CTC method across both baseline models and test sets.
Moreover, our integrated method, incorporating dual monolingual datastores with a gated datastore selection mechanism, demonstrates superior performance in all scenarios. This highlights that the utilization of the bilingual datastore introduces distracting noise during the retrieval process, which adversely affects performance. By implementing the gated mechanism to select a monolingual datastore, we effectively mitigate interference from alternate languages, resulting in precise retrieval and improved performance.
Our approach achieves a relative MER reduction of 4.4\% and 7.4\% on the $TEST$ and $S_{MIX}$ sets respectively for the Conformer baseline. Furthermore, for the Wav2vec2-XLSR baseline, we observe a relative MER reduction of 3.6\% and 5.1\% on the two test sets separately.

Additionally, we compute the Real Time Factor\footnote{\url{https://openvoice-tech.net/index.php/Real-time-factor}} (RTF) for both the Conformer and Wav2vec2-XLSR baselines. Compared with CTC and $k$NN, a slight increase in RTF is evident due to the $k$NN retrieval process. This marginal slowdown is expected and deemed acceptable given the dual retrieval processes involved in our method. These results underscore the effectiveness of our proposed approach.

To further assess the effectiveness of our proposed method, we compare its performance with Whisper \cite{whisper} and Prompting Whisper (PW) \cite{peng2023prompting}, as shown in Table \ref{compare to other}. We reimplement several versions of PW for this comparison. Our method outperforms both Whisper-Small and PW-Small, achieving superior results with fewer parameters and a lower RTF.

\begin{table}[t]
  \caption{MER (\%) of our proposed method based Conformer, in comparison to Whisper and PromptingWhisper (PW) \cite{peng2023prompting}.}
  \label{compare to other}
  \centering
\begin{tabular}{ccccc}
\toprule
Model    &Type &\# Params     & $TEST$   & RTF\\
\midrule
\multirow{2}{*}{Whisper}  &B &74M  & 38.79  & 0.0455      \\
 &S &244M   &27.42  & 0.0912
 \\ \midrule
\multirow{2}{*}{PW \cite{peng2023prompting} }
    &B &74M    &46.38  &0.0458 \\
    &S &244M   &25.70  &0.0923 \\
    \midrule
Ours (Conformer)  & - &123M  & 25.02  & 0.0151 \\
\bottomrule
\end{tabular}
\end{table}

\begin{figure}[!t]
    \centering  
    \subfigure["\begin{CJK}{UTF8}{gbsn}不是关于\end{CJK}
LOVE STORY"]{
        \includegraphics[width=0.45\linewidth]{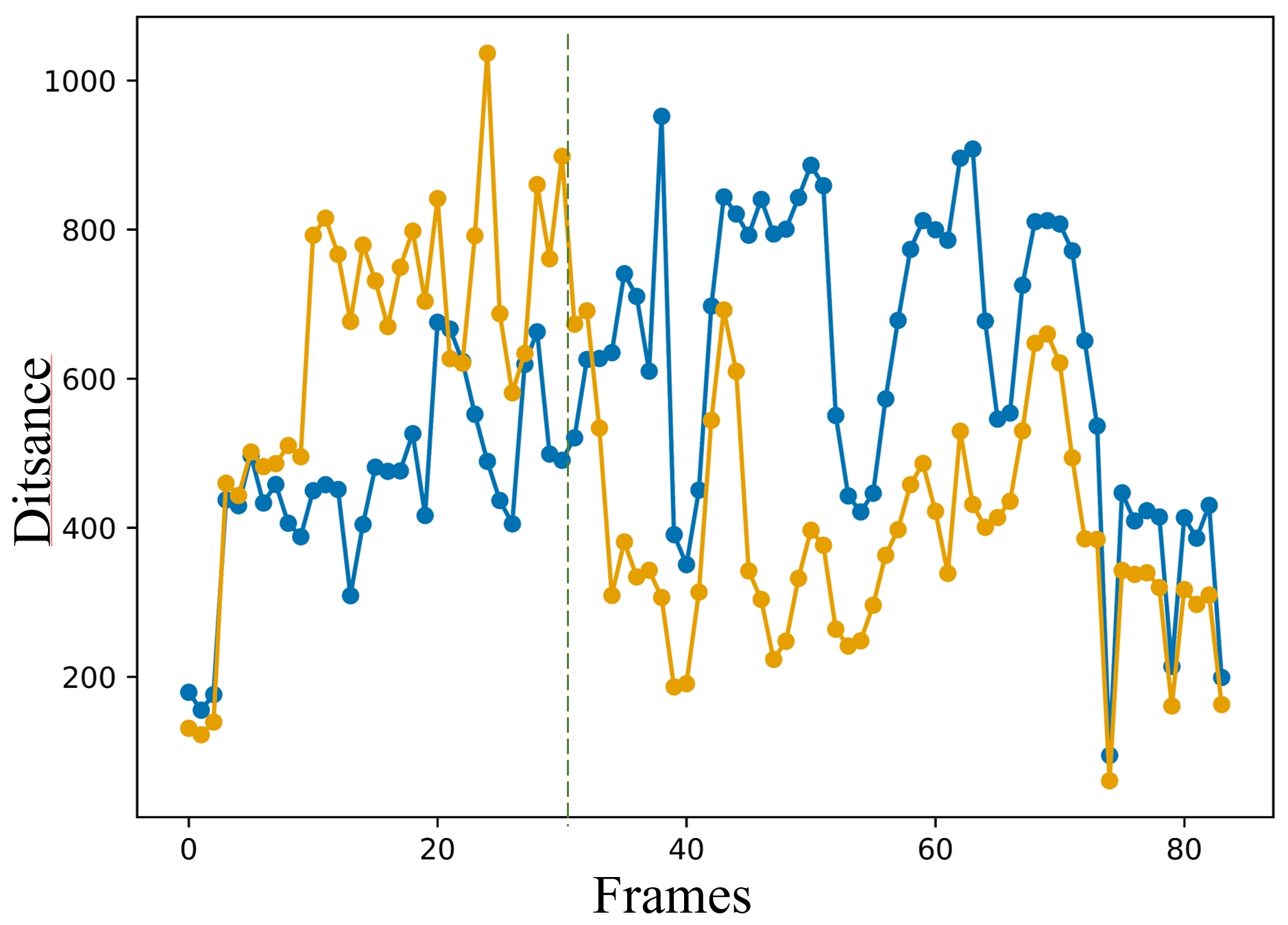}
        \label{visual-1}
    }
    \subfigure["\begin{CJK}{UTF8}{gbsn}就是那种\end{CJK}
  STUDY ENVIRONMENT
  \begin{CJK}{UTF8}{gbsn}特别好\end{CJK}"]{
        \includegraphics[width=0.45\linewidth]{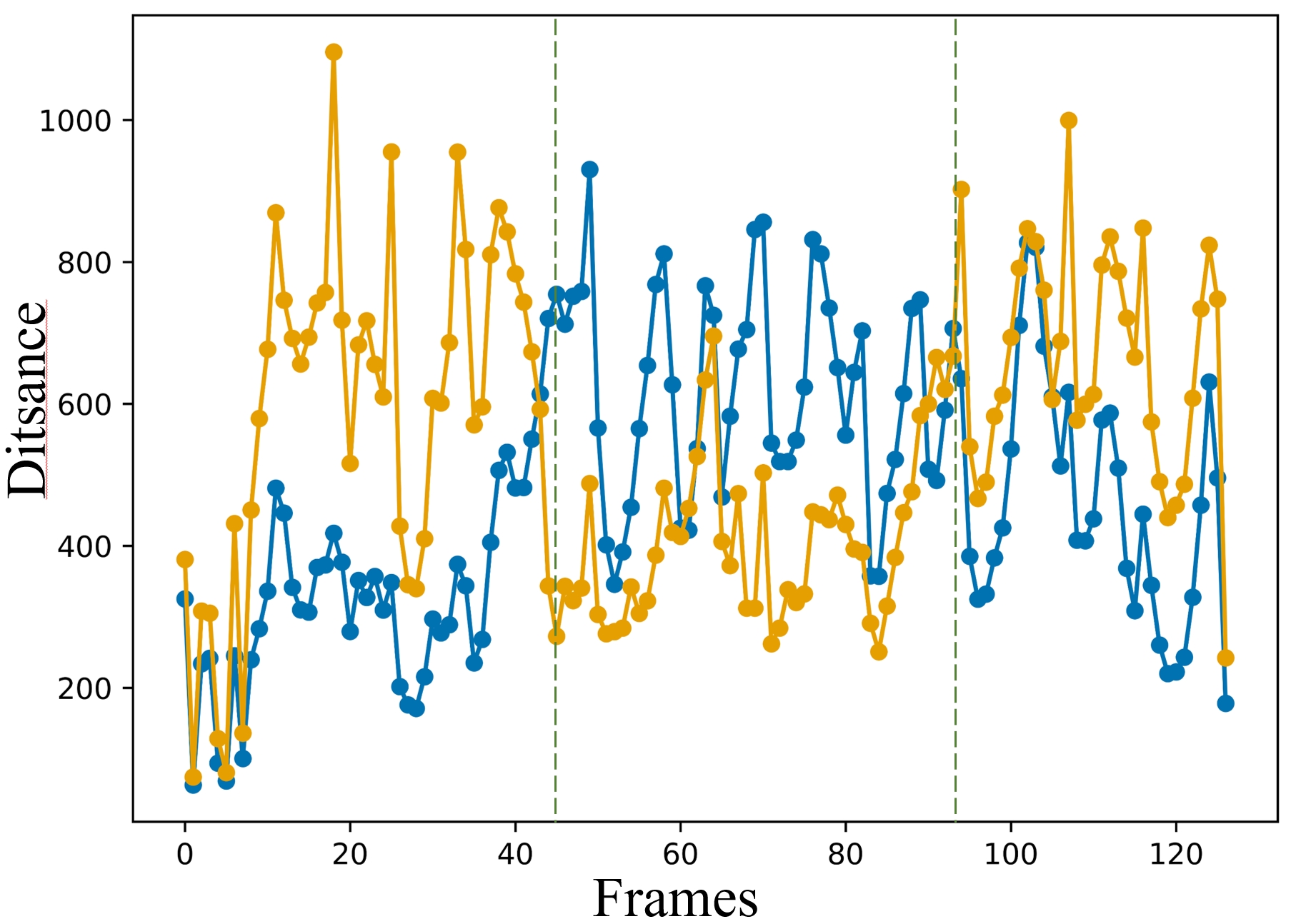}
        \label{visual-2}
    }
    \caption{Visualization of average distances $d_{CN}$ and $d_{EN}$. The green dashed vertical line represents occurrences of CS. The color blue represents Chinese, while orange represents English.}  
    \label{visual}
\end{figure}
In Figure \ref{visual}, we illustrate the average distances, $d_{CN}$ and $d_{EN}$, for each frame. We observe that, during code-switching events where the language shifts, $d_{CN}$ and $d_{EN}$ also switch correspondingly. In both Figure \ref{visual-1} and Figure \ref{visual-2}, the average distance for the current spoken language is consistently lower than that for the alternate language. This observation highlights the effectiveness of our method, demonstrating that average distances can serve as a reliable metric for language identification.

\subsection{Ablation study}

The results of the ablation study conducted on Wav2vec2-XLSR are shown in Table \ref{ablation study}. We report Word Error Rate (WER) for English, Character Error Rate (CER) for Chinese and total MER.
Specifically, we conduct ablations on the utilization of a single datastore (S1), two separate datastores with a gated datastore selection mechanism (S2), and scale temperature $t$ to adjust the alternate language distribution (S3).  
Our findings indicate that $k$NN with a single datastore outperforms the CTC baseline. Moreover, $k$NN with two separate datastores further improves performance by reducing decoding interference from extraneous language noise. 
Furthermore, the scale temperature $t$ has the most significant impact on the overall MER, adjusting the alternate language probability based on the accuracy of the gated datastore selection mechanism. By employing the gated mechanism and scale temperature $t$, we achieve the best performance.

\begin{table}[!t]
  \caption{Ablation study of Wav2vec2-XLSR baseline on $TEST$ and $S_{MIX}$. }
  \label{ablation study}
  \centering
\begin{tabular}{ccccccc}
\toprule
\multirow{2}{*}{Method} & \multicolumn{3}{c}{$TEST$}  & \multicolumn{3}{c}{$S_{MIX}$}\\ 
\cmidrule(r){2-4}  \cmidrule(r){5-7} & CER    & WER     & MER  & CER    & WER     & MER \rule{0pt}{10pt} \\
\midrule

S0      &24.30	&66.83 & 32.65 &23.98 &74.25 &35.22\\
S1   &24.23	&66.19 &32.47 &23.65 &74.08 &34.94\\
S2 &24.04	&66.33 &32.34  &23.49 &73.67 &34.70\\
S3 &\textbf{23.62} &\textbf{63.68} &\textbf{31.48}  &\textbf{22.53} &\textbf{71.25} &\textbf{33.41}\\
\bottomrule
\end{tabular}
\end{table}
\section{Conclusion}

In this paper, we propose a $k$NN-CTC framework that utilizes dual monolingual datastores and implements a gated datastore selection mechanism for zero-shot Chinese-English CS-ASR. Compared to using a bilingual datastore, our method avoids undesirable noise from the alternate language and facilitates the selection of the appropriate monolingual datastore for each frame during decoding. This ensures the explicit injection of language-specific information. Extensive experiments demonstrate the effectiveness of our approach for zero-shot Chinese-English CS-ASR.

\newpage

\bibliographystyle{IEEEbib}
\bibliography{mybib}

\end{document}